\documentclass[conference,transmag]{IEEEtran}
\IEEEoverridecommandlockouts
\usepackage{cite}

\usepackage{amsmath,amssymb,amsfonts}

\usepackage{graphicx}
\usepackage{textcomp}
\usepackage{xcolor}
\usepackage{boldline}
\usepackage{hyperref}
\usepackage{url}
\usepackage{tikz}
\usepackage{multirow}
\usepackage{cite}
\usepackage{multicol}
\usepackage{stfloats}
\usepackage{float}

\usepackage{etoolbox}
\makeatletter
\patchcmd{\@makecaption}
  {\scshape}
  {}
  {}
  {}
\makeatother

\usepackage{algorithm} 
\usepackage{algpseudocode} 

\makeatletter
\newcommand{\multiline}[1]{%
  \begin{tabularx}{\dimexpr\linewidth-\ALG@thistlm}[t]{@{}X@{}}
    #1
  \end{tabularx}
}
\makeatletter
\def\hlinewd#1{%
  \noalign{\ifnum0=`}\fi\hrule \@height #1 \futurelet
  \reserved@a\@xhline}

\def\BibTeX{{\rm B\kern-.05em{\sc i\kern-.025em b}\kern-.08em
    T\kern-.1667em\lower.7ex\hbox{E}\kern-.125emX}}
\begin{document}


\title{CONetV2: Efficient Auto-Channel Size Optimization for CNNs
}
\author{\IEEEauthorblockN{Yi Ru Wang\IEEEauthorrefmark{1}\thanks{$^*$ Equal Contribution}$^*$,
Samir Khaki\IEEEauthorrefmark{1}$^*$,
Weihang Zheng\IEEEauthorrefmark{1}$^*$, 
Mahdi S. Hosseini\IEEEauthorrefmark{2}$^*$,
Konstantinos N. Plataniotis\IEEEauthorrefmark{1}
}
\IEEEauthorblockA{\IEEEauthorrefmark{1}The Edward S. Rogers Sr. Department of Electrical \& Computer Engineering, University of Toronto}
\IEEEauthorblockA{\IEEEauthorrefmark{2}The Department of Electrical and Computer Engineering, University of New Brunswick}

\url{https://github.com/mahdihosseini/CONetV2}
}


\maketitle

\begin{abstract}
Neural Architecture Search (NAS) has been pivotal in finding optimal network configurations for Convolution Neural Networks (CNNs). While many methods explore NAS from a global search-space perspective, the employed optimization schemes typically require heavy computational resources. This work introduces a method that is efficient in computationally constrained environments by examining the micro-search space of channel size. In tackling channel-size optimization, we design an automated algorithm to extract the dependencies within different connected layers of the network. In addition, we introduce the idea of knowledge distillation, which enables preservation of trained weights, admist trials where the channel sizes are changing. Further, since the standard performance indicators (accuracy, loss) fail to capture the performance of individual network components (providing an overall network evaluation), we introduce a novel metric that highly correlates with test accuracy and enables analysis of individual network layers. Combining dependency extraction, metrics, and knowledge distillation, we introduce an efficient searching algorithm, with simulated annealing inspired stochasticity, and demonstrate its effectiveness in finding optimal architectures that outperform baselines by a large margin.


\end{abstract}


\section{Introduction}
Recent advances in Convolution Neural Network (CNN) performance have been associated with the paradigm shift from handcrafted design to Neural Architecture Search (NAS)\cite{lee2020s3nas, wistuba2019survey,baker2017designing, yao2019taking, xie2017genetic}. Focusing on network configuration, NAS inspired state of the art topologies from AlexNet to GoogleNet enabling networks to outperform baseline counterparts \cite{NIPS2012_c399862d, szegedy2014going, real2019regularized}. While NAS helps to find optimal network architectures, there are several challenges that exist in literature. To begin with, there is often an inevitable computational cost associated with searching. The primary bottleneck comes from the search time required to reach optimality. For every searched structure, long training schedules are necessary prior to assessment \cite{zoph2018learning}. As well, standard performance indicators like accuracy and loss only provide a holistic perspective on the network's inference ability, while failing to differentiate performance of individual network components. Further, unique to the case of channel size optimization, the complex structure of many recent CNNs make the optimization of channel sizes difficult due to the inter relationships between the layers. Therefore, motivated by these challenges, there has been explorations in computationally efficient algorithms \cite{liu2018progressive, pham2018efficient, ding2021hrnas,abdelfattah2021zerocost}, metrics to measure performance of networks \cite{lao2020channeldirected, jiang2019fantastic, dziugaite2021search}, and automated NAS. 


\begin{figure}[ht]
    \begin{minipage}{\columnwidth}
    \centerline{\includegraphics[width=\columnwidth]{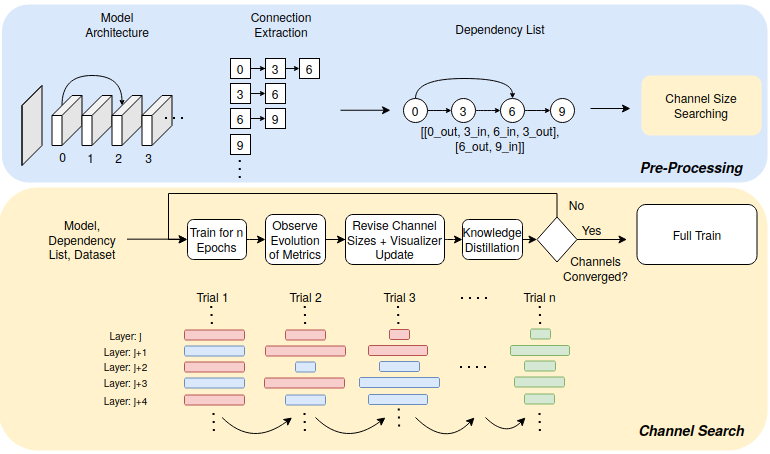}}
    \begin{minipage}[t][4.2em][t]{\linewidth}
    \vspace{-1.2em}
    \caption{
    \textbf{Channel Search Pipeline.} There are two key components to the Channel Search Pipeline: Pre-processing and Channel Searching Modules. The Pre-processing module interprets an initialized model to extract convolution layers and combines this with an adjacency representation to identify layer dependencies in a list. The Channel Searching module illustrates the searching routine for how channel sizes are optimized for a neural architecture.}
    \end{minipage}
    \label{fig:pipeline}
    \end{minipage}
\end{figure}
\begin{figure}[ht]
    \begin{minipage}{\columnwidth}
    \vspace{-1em}
    \centerline{\includegraphics[width=0.35\columnwidth]{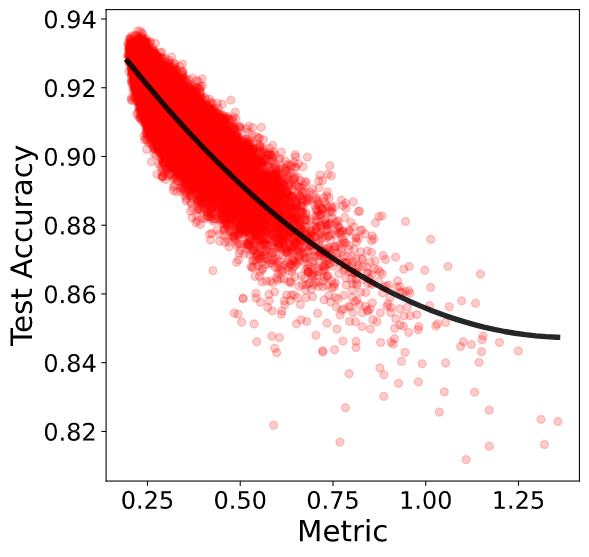}\includegraphics[width=0.365\columnwidth]{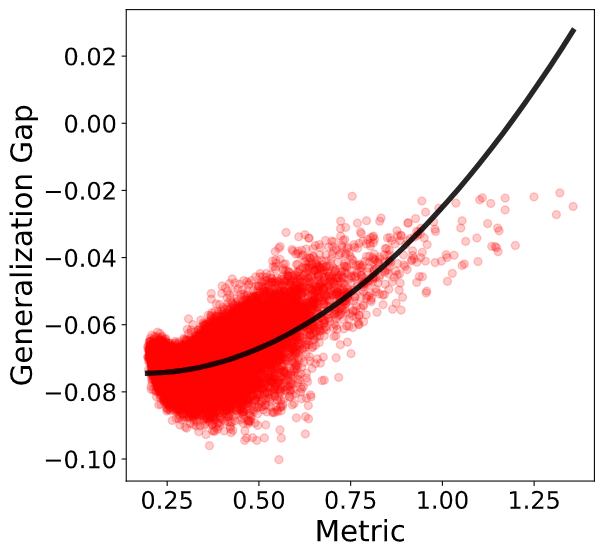}}
    \begin{minipage}[t][0.5em][t]{\linewidth}
    \vspace{-1em}
    \caption{
    \textbf{Metric Correlation} with Test Accuracy and Generalization Gap using trained architectures from the NATS-Bench \cite{Dong_2021}.}
    \end{minipage}
    \label{fig:correlation}
    \end{minipage}
    \vspace{-2.1em}
\end{figure}
Our work is built upon the understanding that central to NAS algorithms is an effective objective function to traverse the search space for an optimal configuration. A metric which has high correlation with test accuracy and generalization gap, as shown in Figure \ref{fig:correlation}, will enable understanding of performance quality of individual layers within the network during architecture search. To tackle the problem of long training schedules, we preserve the weight values of the network between search trials to retain the knowledge and significantly reduce the computation overload required with NAS. Unique to channel size searching is the challenge of inter-dependencies between the channel sizes, we overcome this with an automated dependency extraction algorithm which represents networks in the form of a Directed Acyclic Graph (DAG). 

Our primary contributions are fourfold:
\begin{itemize}
    \item We introduce a metric measure on the performance of individual convolution layers by representing how well each layer is learning. This metric is computed by a norm measure space of the low-rank structure of the weights.
    \item We introduce an automated dependency extraction algorithm which utilizes layer relationships to determine the dependent layers that must be optimized con-currently.
    \item We introduce a new channel search algorithm which scales channel sizes using momentum averaging between scaling trials, and Simulated Annealing (SA) inspired stochasticity to avoid local optimas.
    \item We introduce an effective Knowledge Distillation method, which allows us to transfer weights for convolution layers between trials admist changing channel sizes.
\end{itemize}

\section{Related Works}

 \textbf{Neural Architecture Search (NAS)} aims to find optimal network architectures based on a heuristic optimization of a given task. Works in this field fall into two main categories based on the considered search space: (a) global search-space which optimizes architecture layer types, connections, and hyper-parameters \cite{miikkulainen2017evolving,elsken2017simple,cai2017efficient, jin2019autokeras, Assun_o_2018, byla2019deepswarm,abdelfattah2021zerocost}; and (b) 
constrained-search space which imposes a constraint often on a cell-level \cite{zoph2018learning, zhong2018practical, liu2018progressive,dong2018dppnet} or on an existing structure \cite{cai2018pathlevel}. Our work considers the micro-search-space of channel sizes without the cell level constraint, and finds the optimal channel sizes for each layer based on dependencies in the skeleton network.

\textbf{Weight Transfer} is a key component of NAS algorithms to transfer weights across neural architectures to reduce both time and computational complexity \cite{pmlr-v80-bender18a, cai2017efficient, hinton2015distilling, pham2018efficient, borsos2019transfer, wistuba2019xfernas}. Knowledge distillation, an extension of transfer learning, utilizes an iterative process in which output results from one network are augmented throughout the training procedure of a different network \cite{phuong2021understanding, heo2019comprehensive, zagoruyko2017paying}. Transferring a set of weights from one architecture to a more complex architecture accelerates training of large networks by passing acquired knowledge gain from previous trails \cite{chen2016net2net, romero2015fitnets}. Our method leverages knowledge distillation to facilitate the CNN training procedure.


\textbf{Searching Methods} in CNN architectures presents an NP-Hard problem \cite{targ2016resnet, WU2019119}. NAS techniques have been employed to expedite this process including greedy search and genetic evolution \cite{Vahdat_2020_CVPR, song2021esenas, co-reyes2021evolving}. Greedy search directly optimizes a heuristic but is dependent on the initialization characteristics making it prone to sub-optimal solutions \cite{DBLP:journals/corr/abs-1912-06059}. In contrast, genetic algorithms attempt to reach an absolute optimum. However, their computational cost and requirement for multi-threading make it infeasible for wide-scale experimentation \cite{qu2020evolutionary}. Our method adapts a greedy and simulated annealing approach combining controlled stochasticity with the heuristic value derived from our metric to address both premature convergence and computational cost.

\section{Probing Technique}\label{sec:methods}
The proposed method uses channel size optimization to explore NAS through the pipeline in Figure \ref{fig:pipeline}. 
In Section \ref{subsec:dependency}, we describe the pre-processing module of Figure \ref{fig:pipeline} featuring the conversion from model architecture to a DAG and subsequent dependency list. In Section \ref{subsec:metrics}, we provide the motivation behind the metric, Quality Condition (QC). 

\subsection{Preliminaries}
Central to our new metric and knowledge distillation technique is the low-rank factorization and Singular Value Decomposition (SVD). Let the weight tensor of a single convolution layer be denoted as $\mathbf{W} \in \mathbb{R}^{N_1 \times N_2 \times N_3 \times N_4}$, where $N_1$ and $N_2$ denote kernel sizes, and $N_3$ and $N_4$ denote the input and output channel sizes, respectively. We first unfold this to a two-dimensional matrix along a given dimension $d$.
\begin{equation}
    \mathbf{W}_{\text{4D}} [\text{Tensor}] \xrightarrow[\text{Unfold}]{\text{Mode-d}} \mathbf{W}_{\text{2D}} [\text{Matrix}].
\end{equation}
Because the 2D weight matrices, $\mathbf{W}$, are perturbed by randomness, similar to \cite{hosseini2020adas}, we decouple the meaningful weight from random perturbation using the low-rank factorization
\begin{equation}
    \mathbf{W} \xrightarrow{factorize} \widehat{\mathbf{W}} + \mathbf{E}.
\end{equation}
The unfolded weight matrix is then decomposed via SVD $\widehat{\mathbf{W}} = \widehat{\mathbf{U}} \widehat{\mathbf{\Sigma}} \widehat{\mathbf{V}}^T$, where $\widehat{\mathbf{\Sigma}} = diag\{\sigma_1, \sigma_2, ..., \sigma_{N'} \}$ with $N' = rank \ \widehat{\mathbf{W}}$. For more information, please refer to \cite{hosseini2020adas} and reference therein. 

\algnewcommand{\IIf}[1]{\State\algorithmicif\ #1\ \algorithmicthen}
\algnewcommand{\EndIIf}{\unskip\ \algorithmicend\ \algorithmicif}
\subsection{Metrics}\label{subsec:metrics}
Borrowing rank measure and condition number from \cite{hosseini2020adas}, we introduce a new metric, dubbed Quality Condition (QC), which provides an aggregated value that encompasses channel capacity and numerical stability of the convolution layer. $\mathcal{R}(\widehat{\mathbf{W}})$ represents the ratio between number of non-zero low-rank singular values and the given input or output channel size. This is a indicator for the encoding capacity of the convolution layer's channel size configuration. A large ratio means that the layer is saturated, while a small ratio means that the layer's weights are under-utilized. The denominator provides a normalized condition, which measures the sensitivity to input perturbations, and takes a value between 0 and 1.
\vspace{-0.4em}
\begin{align}
    QC = \arctan(\frac{r}{1-1/\kappa})~~\text{where}
\end{align}
\begin{align}
    \mathcal{R}(\widehat{\mathbf{W}}) = {N'}/{N}~~\text{and}~~\kappa(\widehat{\mathbf{W}}) = {\sigma_1(\widehat{\mathbf{W}})}/{\sigma_{N'}(\widehat{\mathbf{W}})}
\end{align}

\subsection{Channel Dependency}
\label{subsec:dependency}

Algorithm \ref{Algo:dependency} introduces a domain agnostic method of determining channel dependencies across convolution layers in any neural architecture given its corresponding DAG. We define a channel dependency as a relationship between the input and output of two or more convolution layers and begin by establishing each neural architecture as a combination of unique components: \textit{Layers}.

\begin{figure*}
\caption{\textbf{Algorithms.} Introducing our novel approach to automatic channel dependency extraction in Algorithm \ref{Algo:dependency} and new searching methods in Algorithm \ref{Algo:Greedy} and \ref{Algo:SA}. The following are key parameter definitions used throughout the paper and below algorithms: $\mathcal{D}$ denotes the list of network dependencies,  $T$ denotes the number of trials, $\mathcal{M}$ denotes the layer based metric result, $m^t_{d_{i}}$ denotes the temporal momentum value at trial $t$ and dependency list index $d_{i}$, $\gamma$ denotes the momentum scaling factor, $\mathcal{S}$ denotes the channel sizes. For Algorithm \ref{Algo:SA}, $\alpha$ denotes a scaling coefficient for the acceptance function, $\text{Temp}$ denotes the temperature scaling,  $\zeta$ denotes the value of the accepting function.}
\vspace{-0.9em}
\algrenewcommand\algorithmicindent{0.75em}%
\begin{minipage}[t]{0.29\textwidth}
\begin{algorithm}[H]
	\caption{Channel Dependency}
	\hspace*{\algorithmicindent} \textbf{Input:} $DAG$\\
    \hspace*{\algorithmicindent} \textbf{Output:} Dependency  $\mathcal{D} $
    \setlength{\columnsep}{1.9em}
    
	\begin{algorithmic}[1]
	\scriptsize{}
	    \State $Visited$, $\mathcal{D}$ $\leftarrow$ [], []
	    \For{$l_n$ in $DAG$}
	        \State $d$, $BackProp$ $\leftarrow$ [], Empty
    	    \For{$l_{n+1}$ \textbf{in} \Call{Next}{$l_n$}}
    	        
                \If{$l_{n+1}$ \textbf{in} $Visited$}
                    \State $BackProp \leftarrow$ ID(\Call{$\mathcal{D}$}{$l_n$})
                \EndIf
                \If{$l_{n+1}$ \textbf{not\ in} $Visited$}
                    \State $Visited$, $d$ $\leftarrow$ $l_{n+1}$
                \EndIf
                
            \EndFor
            \If{$l_n$ \textbf{in} $Visited$}
                \State $BackProp \leftarrow$ ID(\Call{$\mathcal{D}$}{$l_n$})
            \EndIf
            \If{$l_n$ \textbf{not\ in} $Visited$}
                \State $Visited$, $d$ $\leftarrow$ $l_n$
            \EndIf
            \If{$BackProp$ $\neq$ $\emptyset$} 
                \State \Call{$\mathcal{D}$}{$BackProp$} $\leftarrow$ $d$\EndIf
            \State \textbf{else:} $\mathcal{D}$ $\leftarrow$ $\mathcal{D}$ + $d$
        \EndFor
	\end{algorithmic}
	\label{Algo:dependency}
\end{algorithm}
\end{minipage}
\begin{minipage}[t]{0.34\textwidth}
\begin{algorithm}[H]
	\caption{Greedy Algorithm}
	\hspace*{\algorithmicindent} \textbf{Input:} $\mathcal{S}^1$, Trials $T$ \\
    \hspace*{\algorithmicindent} \textbf{Output:} \textbf{$\mathcal{S}^{optim}$}
	\begin{algorithmic}[1]
	\scriptsize{}
	    \For{$t$ \textbf{in} $1:T$}
	        \If{t = 1}
	            \State $m_{d_i, :}^{0} \leftarrow 0$
	        \EndIf
	        \For{$d_i, d$ \textbf{$in$} $\mathcal{D}$}
	            \For{$l_i, l$ in $d$}
	                \State $\mathcal{M}_{d_i, l_i} \leftarrow$ \Call{ComputeMetric}{$l$}
	            \EndFor
    	        \State $m_{d_i}^t \leftarrow$ $\gamma * m_{d_i}^{t-1} +  \frac{\sum_{l_i \in |d|} M_{d_i,l_i}}{|d|}$
    	        
    	        \State $\Delta m_{d_i} \leftarrow$ $m_{d_i}^t - m_{d_i}^{t-1}$
    	        
    	        
    	        \State $\Delta \mathcal{S}_{d_i}^t \leftarrow$ \Call{Clip}{$1 + \Delta m_{d_i}$, [0.5, 2]}
	        
	            \State $\mathcal{S}_{d_i}^t \leftarrow$ $\mathcal{S}_{d_i}^{t-1}*\Delta \mathcal{S}_{d_i}^t$
	            \State{\textit{Train model for $e$ epochs}}
	            \If{$m_{d_i, :}^{t} > m_{d_i, :}^{Best}$}
	                \State $m_{d_i, :}^{Best} \leftarrow m_{d_i, :}^{t}$
	            \EndIf
	       \EndFor
	    \EndFor
	\end{algorithmic}
	\label{Algo:Greedy}
\end{algorithm}
\end{minipage}
\begin{minipage}[t]{0.34\textwidth}
\begin{algorithm}[H]
	\caption{Simulated Annealing}
	\hspace*{\algorithmicindent} \textbf{Input:} $\mathcal{S}^1$, Trials $T$, $\alpha$\\
    \hspace*{\algorithmicindent} \textbf{Output:} \textbf{$\mathcal{S}^{optim}$}
    
    \setlength{\columnsep}{1.9em}
	\begin{algorithmic}[1]
	\scriptsize{}
	    \For{$t$ \textbf{in} $1:T$} \hspace{0.1cm} \text{Temp} $\leftarrow$ $\alpha * \frac{(T-t)}{T}$
	        \If{t = 1}
	            \State $m_{d_i, :}^{0} \leftarrow 0$
	        \EndIf
	        
	        \For{$d_i, d$ \textbf{$in$} $\mathcal{D}$}
	            \For{$l_i, l$ in $d$}
	                \State $\mathcal{M}_{d_i, l_i} \leftarrow$ \Call{ComputeMetric}{$l$}
	            \EndFor
	            
    	        \State $m_{d_i}^t \leftarrow$ $\gamma * m_{d_i}^{t-1} +  \frac{\sum_{l_i \in |d|} M_{d_i,l_i}}{|d|}$
    	        
    	        \State $\Delta m_{d_i} \leftarrow$ $m_{d_i}^t - m_{d_i}^{t-1}$
    	        \State $\zeta = e^{-1/(\alpha*\Delta m_{d_i}* \text{Temp})} $
    	       \If{$x \epsilon [0,1] < \zeta $} 
    	            \State $\Delta m_{d_i} \leftarrow \Delta m_{d_i} + \zeta $
    	       \EndIf
    	        
    	        
    	        \State $\Delta \mathcal{S}_{d_i}^t \leftarrow$ \Call{Clip}{$1 + \Delta m_{d_i}$, [0.5, 2]}
	        
	            \State $\mathcal{S}_{d_i}^t \leftarrow$ $\mathcal{S}_{d_i}^{t-1}*\Delta \mathcal{S}_{d_i}^t$
    	        
	            \State{\textit{Train model for $e$ epochs}}
	            
	            \If{$m_{d_i, :}^{t} > m_{d_i, :}^{Best}$} 
	                \State $m_{d_i, :}^{Best} \leftarrow m_{d_i, :}^{t}$
	            \EndIf
	       \EndFor
	    \EndFor
	\end{algorithmic}
	\label{Algo:SA}
\end{algorithm}
\end{minipage}
\vspace{-1.8em}
\end{figure*}

For the purpose of channel dependency, a \textit{layer} is considered to be any element in the neural architecture with a 4D-Weight tensor -- most commonly convolution and de-convolution layers. A \textit{layer} is made of two elements: The \textit{Input} and \textit{Output Channel}.

\begin{figure}[H]
\begin{minipage}{\columnwidth}
    \vspace{-1.4em}
    \centerline{\includegraphics[width=0.6\columnwidth]{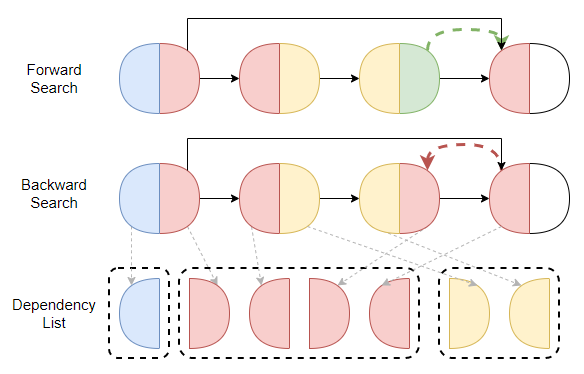}}
     \begin{minipage}[t][4.2em][t]{\linewidth}
    \vspace{-1.2em}
    \caption{\textbf{Dependency Extraction.} We forward traverse the network until reaching a layer with a pre-defined dependency. This triggers a back-progagation assignment of dependencies, assigning the current iteration of layers to the pre-defined dependency group. The resultant dependencies represent the channel sizes that must be joint-optimized.}
    \end{minipage}
    \label{fig:dependency}
    \end{minipage}
     \vspace{-1.2em}
\end{figure}

Using the basis of these multi-channeled layers, we establish Algorithm \ref{Algo:dependency} to identify dependencies across every layer's channels in the model architecture. When traversing the DAG representation of the model, the algorithm conducts forward propagation for dependency construction. The output channel size and the input channel size of the connected layers are generally placed within the same dependency group. Network channels which have been assigned dependencies are tracked. Upon encountering an element which has previously been assigned to a dependency group, back propagation is triggered to combine dependency groups. A visual depiction of the dependency extraction process is shown in Figure \ref{fig:dependency}. 

\section{Channel Searching}
The resultant dependency list contains two categories of channels: \textit{independent} and \textit{dependent}. Independent channels are considered to be any channel of a layer whose corresponding channel size can be altered without affecting any preceding layer's channel sizes. Conversely, dependent channels are defined as any channel whose size is constrained to match any preceding layer's channel size. Combining dependencies with calculated metrics, we introduce \ref{subsec:searching_alg} which discusses our novel searching algorithms. To preserve the qualities of a layer as its size changes between search trials, we introduce knowledge distillation in \ref{subsec:KD}. We show that an efficient approach to the scaling of channel sizes yields an improvement in the overall network performance.
\subsection{Optimization Problem}
We define the optimization problem as follows:
\begin{equation}\label{eq_opt_problem}
\begin{gathered}
\max_{t \in T}\sum_{d_i \in |\mathcal{D}|}m_{d_i}^t\\
   \text{s.t.}~~m_{d_i}^t = \gamma * m_{d_i}^{t-1} +  \frac{\sum_{l_i \in |d|} M_{d_i,l_i}}{|d|},
\end{gathered}
\end{equation}
where, $m_{d_i}^t$ is the average momentum metric, $t \in 1 ... T$ refers to the trial number, $d$ denotes the dependency list at dependency index $d_i$, $\gamma$ represents the momentum scaling factor, and $M_{d_i,l_i}$ represents the QC Metric computed on layer $l_i$ within the dependency list $d_i$. The goal is to maximize the objective in (\ref{eq_opt_problem}) to achieve the optimal set of channel sizes for the given architecture.


\subsection{Searching Algorithm} \label{subsec:searching_alg}
We recast the solution to the optimization problem in (\ref{eq_opt_problem}) by Algorithms \ref{Algo:Greedy} and \ref{Algo:SA}. We present the searching process for channel size optimization, using the greedy and simulated annealing optimization schemes, respectively. We denote $m^{t-1}$ as the previous cumulative metric, $\mathcal{D}$ as the dependencies, and $t$ as the trial number.

\textbf{Metric Computation} involves applying the QC metric as highlighted in Section \ref{subsec:metrics}. This includes flattening the weight tensor of the respective layer $\mathbf{W}_{l_i}$ along the dimension $d$, based on whether it is an input or output channel. We then apply SVD on the resulting 2D-matrix, and use the resulting bases and singular values for metric computation.

\textbf{Momentum} is used to accumulate metric information of previous trials with the current trial. $\gamma$ regulates the ratio of historical to present information during accumulation. Momentum is applied to smooth out the metric difference between trials, and stabilizes the algorithm from sudden changes in metric value occurring between trials.

\textbf{Greedy Search} is introduced in Algorithm \ref{Algo:Greedy} as a method of channel size optimization through selection of the best locally determined channel size based on the QC metric. For every trial, Algorithm \ref{Algo:Greedy} probes each \textit{layer} and computes the metric. This metric is then aggregated across unique dependency lists and used to guide the direction scaling of the respective channel sizes. By combining the metrics value with the use of dependency list aggregation, the algorithm is able to change every unique channel size without visiting every \textit{layer}.  

\textbf{Simulated Annealing} is an extension of Greedy Search that adds a degree of stochasticity to the channel scaling process. Introduced in Algorithm \ref{Algo:SA}, the main components are the accepting function, $\zeta$, and the temperature scaling, \text{Temp}. The accepting function was created to ensure an uneven distribution of values $\in [0,1]$ such that the probability of a random number $\epsilon \in [0,1]$ being less than $\zeta$ would naturally decrease over trials. The temperature scaling was incorporated to further modulate the probability of randomness by amplifying the rate of channel size exploration in the earlier trials. In the cases of acceptance ($\epsilon \in [0,1] \leq \zeta $), the value of the temperature dependent accepting function was added to the metric causing a greater scaling magnitude. This leads to an increase in the likelihood of escaping any local optima. As trials progress, both $\text{Temp}$ and $\zeta$ work to reduce the probability of randomness thus stabilizing the algorithm towards convergence. Due to the randomness of exploration, the algorithm keeps track of the heuristically determined best channel sizes throughout all trials to ensure all explored options were considered when determining the final channel size. 



\subsection{Knowledge Distillation}\label{subsec:KD}
We introduce knowledge distillation to reduce the computational cost associated with training neural networks from random initialization for each trial. Between successive searching trials, the weight tensors will be initialized based on converged weights from the previous trial. Between trials, one of three cases can arise for distillation of channel dimension $d$ of a convolution layer, where $d \in \{3,4\}$ for input or output channel, respectively. The channel size can increase, decrease, or remain constant, from the size of one trial $N_d$, to the size of the succeeding trial $N_d'$. When $N_d = N_d'$, the weight tensor can be replaced without any modification along the given dimension $d$. Below, we provide derivations for cases involving channel expansion or shrinkage.

\textbf{Channel Expansion}. We denote $D = N_d' - N_d$, where $N_d < N_d'$ . To increase the channel size, we apply the reflection operation to the unfolded 2D Matrix. This means that we will concatenate the matrix with a reflection of its last $D$ rows $\mathbf{W}' \leftarrow$ $[\mathbf{W}; \mathbf{W}_{N_d:-1:2N_d-N_d',:}]$.

\textbf{Channel Shrinkage}. We denote $D = N_d - N_d'$ , where $N_d' < N_d$. Taking the SVD of the unfolded 2D Matrix, $\mathbf{W} = U\Sigma V^T$, we select the top $N_d'$ rows of $U$, and the top left $N_d' \times N_d'$ region of $\Sigma$, then multiply the new decomposition to get our row reduced matrix. Note that we select the top region to preserve the optimal bases of the weight matrix.





\begin{minipage}{0.48\textwidth}
\begin{algorithm}[H]
    \caption{Knowledge Distillation}\label{algorithm}
    \footnotesize
    \hspace*{\algorithmicindent} \textbf{Input:} $\displaystyle \mathbf{W}^{l}_{j-1}$, $n_d'$\\
    \hspace*{\algorithmicindent} \textbf{Output:} $\displaystyle \mathbf{W}^{l}_{j}$
	\begin{algorithmic}[1]
	\scriptsize{}
	    \State $n_w$, $n_h$, $n_i$, $n_o$ $\leftarrow$ \Call{shape}{$\displaystyle \mathbf{W}^{l}_{j-1}$}
	    \State $\displaystyle \mathbf{W}_{j-1}^{l}$[Tensor-4D] $\xrightarrow{\text{unfold}}$ $\mathbf{W}$[Matrix-2D]
	    \If {$n_d>n_d'$}
	        \State $\mathbf{U}, \mathbf{\Lambda}, \mathbf{V} =$ SVD($\mathbf{W}$)
	        
	        \State $U',\hspace{0.1cm} \Lambda',  \leftarrow \mathbf{U}_{1:n_d',1:n_d'},\hspace{0.1cm} \mathbf{\Lambda}_{1:n_d',:}$
            \State $\mathbf{W}' \leftarrow \mathbf{U}'\mathbf{\Lambda}' \mathbf{V}^T$
	    \EndIf
	    \IIf {$n_d<n_d'$} {$\mathbf{W}' \leftarrow$ {[$\mathbf{W}; \mathbf{W}_{n_d:-1:2n_d-n_d',:}$]}}\EndIIf
        \State $\displaystyle \mathbf{W}^{l}_{j}$[Tensor-4D] $\xleftarrow{\text{reshape}}$ $\mathbf{W}'$[Matrix-2D]
	\end{algorithmic} 
	\label{algo:KD}
\end{algorithm}
\end{minipage}
\hfill

\section{Experiments}
\begin{figure*}[t]

\centering
\begin{minipage}[t]{0.48\textwidth}
  \centering
  \includegraphics[width=\textwidth, height=0.4\textwidth]{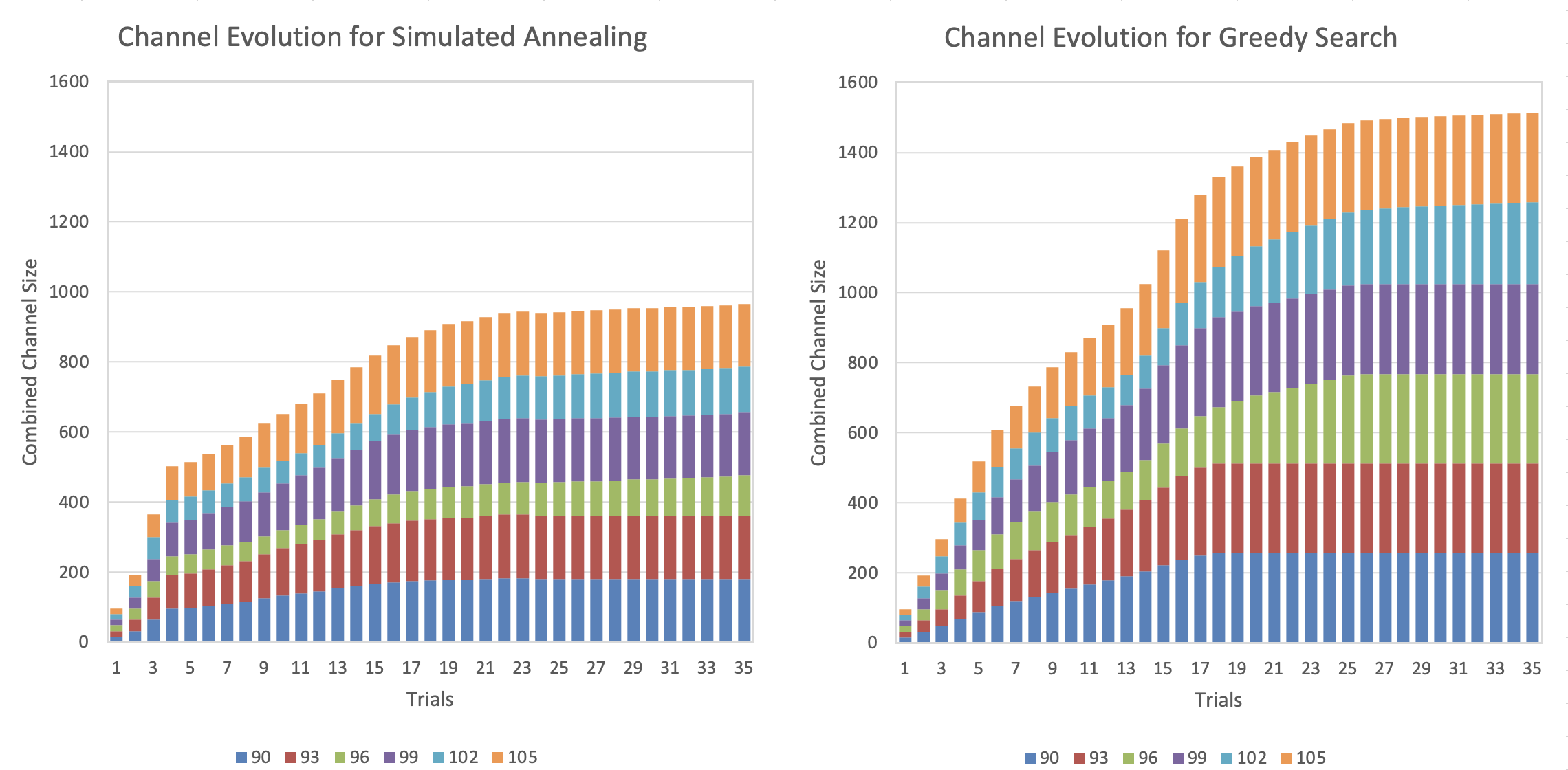}
 \begin{minipage}[t][6em][t]{\linewidth}
    \vspace{-1em}
  \caption{Channel evolution of the final 6 \textit{layers} (90,93,96,99,102,105)
  in ResNet34 over the searching period of 35 trials for both Simulated Annealing (left) and Greedy Search (right). The channel convergence data was from experiments of SA and Greedy which achieved comparable final accuracy ($\pm$ 0.1). The SA converges faster and to a lower overall channel size compared to Greedy Search whilst still retaining a comparable accuracy thus demonstrating it's efficacy as a robust and efficient algorithm.}
  \end{minipage}
 
  \label{fig:ChannelEvolution}
\end{minipage}%
\hspace{0.5em}
\begin{minipage}[t]{0.48\textwidth}
 
  \centering
  \includegraphics[width=0.49\textwidth]{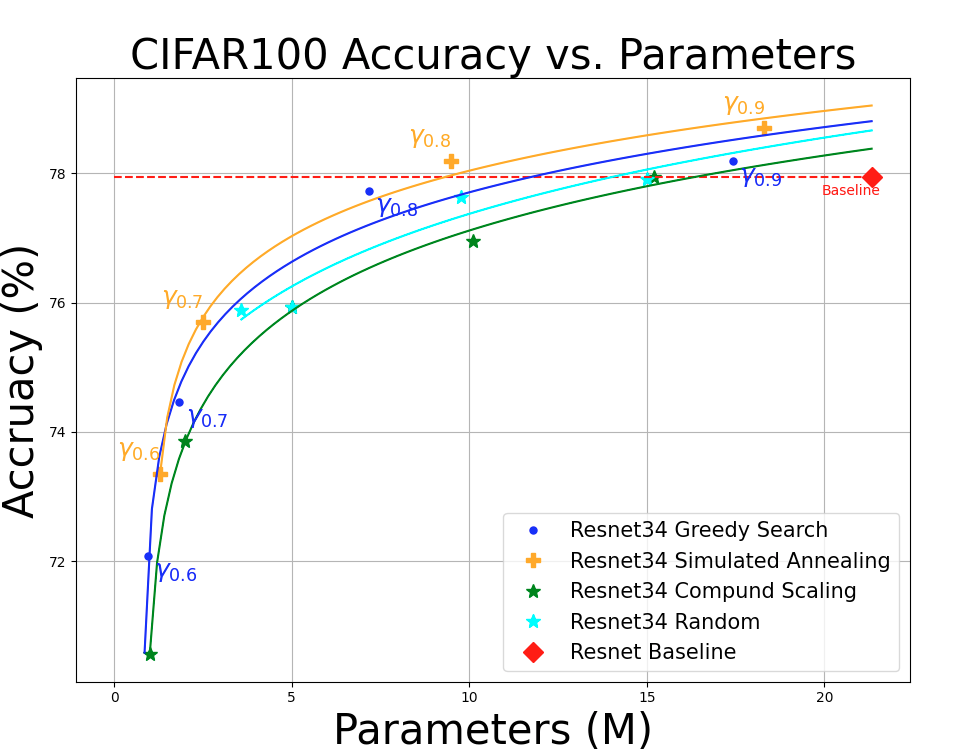}
  \includegraphics[width=0.49\textwidth]{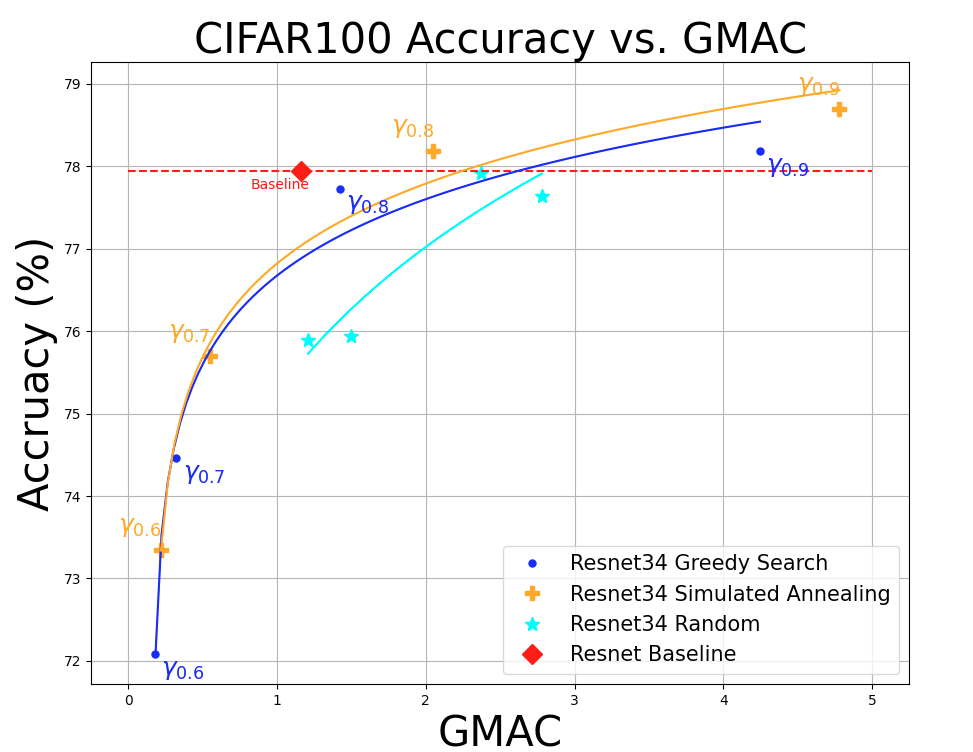}
 \begin{minipage}[t][6em][t]{\linewidth}
    \vspace{-1em}
  \caption{Shows the Accuracy vs. Parameters (left) and Accuracy vs. GMAC (right) of the Resnet34 model using Algorithms \ref{Algo:Greedy} and \ref{Algo:SA}. For comparative purposes, the compound scaling \cite{tan2020efficientnet}, random scaling, and baseline are also shown. As illustrated above, the Simulated Annealing and Greedy Search Algorithms outperform other competitive strategies in both parameters and GMAC against accuracy thus validating the efficacy of presented metric and Algorithms \ref{Algo:Greedy} and \ref{Algo:SA}.}
  \end{minipage}
  
  \label{fig:ResnetAccParam}
\end{minipage}
\end{figure*}
\subsection{Implementation Details}
\textbf{ResNet34 \cite{he2015deep}.} Prior to searching, output channel sizes of convolution layers are initialized to 16. We search for a total of 35 trials, with two epochs per trial. During searching, we use an initial learning rate of 0.03 on the AdaS Optimizer, with momentum 0.9 and weight decay 0.0005 \cite{hosseini2020adas}, and $\alpha=5$ for SA experiments. After searching, we conduct full train for 250 epochs. We use SGD Optimizer with initial learning rate 0.1, momentum 0.9, weight decay 0.0005 and the StepLR scheduler with step size 25, and gamma 0.5 \cite{10.1214/aoms/1177729586}.

\textbf{DARTS7 \cite{liu2019darts}.} The conditions for DARTS7 follow that of ResNet34 with a few exceptions. During searching, we use an initial learning rate of 0.175 on the AdaS Optimizer, with momentum 0.9 and weight decay 0.0003 \cite{hosseini2020adas}, and $\alpha=5$ for SA experiments. After searching, we conduct full train for 250 epochs. We use SGD Optimizer with initial learning rate 0.175, momentum 0.9, weight decay 0.0003, and the StepLR scheduler with step size 25 and gamma 0.5 \cite{10.1214/aoms/1177729586}. We \textbf{do not} apply cutout during experimentation.



    

In addition to the above hyper-parameters, the neural architectures are run through the same experimental testing conditions. Each initialized model architecture is fed through the automated pipeline which extracts channel dependencies across all \textit{layers}, using Algorithm \ref{Algo:dependency}, to be used channel modulation. Greedy or Simulated Annealing Channel Search is then applied to the Neural Architecture. Knowledge distillation is then used to transfer learned parameters for the next trial - see Algorithm \ref{Algo:Greedy}, \ref{Algo:SA}, \ref{algo:KD}. After the 35 trials, the best selected channel sizes are used to re-initialize a new model for full train. All experimentation was completed with CIFAR10 and CIFAR100 \cite{cifar}. Results are highlighted in Table \ref{table_results}.

\begin{table*}[htp]

\setlength\tabcolsep{1pt}
\centering
\begin{minipage}[t][6.5em][t]{\linewidth}
    \vspace{-1.2em}
    \caption{Comparison of performance of \textbf{Greedy Search} (Algorithm \ref{Algo:Greedy}) and \textbf{Simulated Annealing} (Algorithm \ref{Algo:SA}) on ResNet34 and DARTS$7$, searched on CIFAR10/100 with different momentum thresholds. Note experimental tests follow designation $Y(\gamma_{n})$, where $Y$ is the applied Algorithm, and $n$ in $\gamma_{n}$ denotes the gamma value for that test. As shown, both Greedy and Simulated Annealing are capable of finding optimal architectures that can outperform the baseline. SA generally yields more efficient architectures (less parameters for the same accuracy) compared to Greedy. Note that all experiments were run at least 2 times to ensure reproducibility.}
   \end{minipage}

    \label{table_results}
    
	\scriptsize{
    \begin{tabular}{c|cc|cc||cc|cc}
	\hlinewd{1.3pt}
    \multicolumn{1}{c|}{}&\multicolumn{4}{c||}{CIFAR10} &
    \multicolumn{4}{c}{CIFAR100}\\
	\hlinewd{1.3pt}
    \multirow{2}{*}{\textbf{Architecture}}&\textbf{Top-1}& \textbf{Params}&\textbf{Search Cost} &\multirow{2}{*}{\textbf{GMAC}}&\textbf{Top-1}& \textbf{Params}&\textbf{Search Cost} &\multirow{2}{*}{\textbf{GMAC}}\\
    &\textbf{(\%)}&\textbf{(M)}&\textbf{(GPU-days)}&&\textbf{(\%)}&\textbf{(M)}&\textbf{(GPU-days)}
    \\
    \hline
    \hline
    ResNet34(baseline)\cite{he2015deep}
    &$95.53_{\pm0.10}$&$21.282$&Manual&$1.162$
    &$77.94_{\pm0.03}$&$21.328$&Manual&$1.162$
    
    \\
    
    Greedy($\gamma_0$)
    &$91.41_{\pm0.57}${\color{red}($-4.12$)}&$0.144$&$0.012$&$0.038$
    &$63.47_{\pm0.45}$ {\color{red}($-14.47$)}&$0.170$&$0.012$&$0.037$
    
    \\
    
    Greedy($\gamma_{0.5}$)
    &$93.51_{\pm0.35}${\color{red}($-2.02$)}&$0.374$&$0.012$&$0.108$
    &$68.58_{\pm0.34}$ {\color{red}($-9.36$)}&$0.367$&$0.012$&$0.067$
    
    \\
    
    Greedy($\gamma_{0.6}$)
    &$93.95_{\pm0.31}$ {\color{red}($-1.58$)}&$0.556$&$0.013$&$0.178$
    &$72.08_{\pm0.70}$ {\color{red}($-5.86$)}&$0.952$&$0.013$&$0.183$
    
    \\
    
    Greedy($\gamma_{0.7}$)
    &$93.97_{\pm0.50}${\color{red}($-1.56$)}&$0.618$&$0.15$&$0.162$
    &$74.46_{\pm0.22}$ {\color{red}($-3.48$)}&$1.812$&$0.15$&$0.322$
    
    \\
    
    Greedy($\gamma_{0.8}$)
    &$95.61_{\pm0.08}$ {\color{green}($+0.08$)}&$4.597$&$0.021$&$1.242$
    &$77.73_{\pm0.35}$ {\color{red}($-0.21$)}&$7.166$&$0.022$&$1.422$
    
    \\
    
    Greedy($\gamma_{0.9}$)
    &$\mathbf{95.84_{\pm0.26}}$ {\color{green}($+0.31$)}&$13.12$&$0.037$&$3.98$
    &$\mathbf{78.19_{\pm0.38}}$ {\color{green}($+0.25$)}&$17.43$&$0.037$&$4.246$
    
    \\\hline

    SA($\gamma_0$)&
    $92.87_{\pm0.29}${\color{red}($-2.66$)}&$0.276$&$0.012$&$0.076$
    &$68.32_{\pm0.12}$ {\color{red}($-9.62$)}&$0.43$&$0.012$&$0.084$
    
    \\
    
    SA($\gamma_{0.5}$)
    &$93.80_{\pm0.51}$ {\color{red}($-1.73$)}&$0.484$&$0.012$&$0.128$
    &$70.40_{\pm0.48}$ {\color{red}($-7.54$)}&$0.673$&$0.012$&$0.120$
   
    \\
    
    SA($\gamma_{0.6}$)
    &$94.57_{\pm0.26}$ {\color{red}($-0.96$)}&$0.956$&$0.014$&$0.254$
    &$73.35_{\pm0.49}$ {\color{red}($-4.59$)}&$1.288$&$0.014$&$0.220$
    
    \\
    
    SA($\gamma_{0.7}$)
    &$95.36_{\pm0.15}$ {\color{red}($-0.17$)}&$2.523$&$0.25$&$0.676$
    &$75.70_{\pm0.49}$ {\color{red}($-2.24$)}&$2.502$&$0.25$&$0.548$
    
    \\
    
    SA($\gamma_{0.8}$)
    &$95.71_{\pm0.01}$ {\color{green}($+0.18$)}&$4.995$&$0.036$&$1.245$
    &$\mathbf{78.19_{\pm0.32}}$ {\color{green}($+0.25$)}&$9.472$&$0.036$&$2.05$
    
    \\
    
    SA($\gamma_{0.9}$)
    &$\mathbf{95.98_{\pm0.06}}$ {\color{green}($+0.45$)}&$14.09$&$0.04$&$3.47$
    &$\mathbf{78.64_{\pm0.09}}$ {\color{green}($+0.70$)}&$18.31$&$0.04$&$4.78$
    
    \\\hline

    DARTS7(baseline)\cite{liu2019darts}
    &$93.37_{\pm0.78}$&$0.223$&Manual&$0.041$
    &$71.15_{\pm0.04}$&$0.246$&Manual&$0.041$
    
    \\
    
    Greedy($\gamma_0$)
    &$91.68_{\pm0.37}${\color{red}($-1.69$)}&$0.092$&$0.033$&$0.048$
    &$68.11_{\pm0.24}$ {\color{red}($-3.21$)}&$0.115$&$0.031$&$0.053$
    
    \\
    
    Greedy($\gamma_{0.5}$)
    &$93.12_{\pm0.19}${\color{red}($-0.25$)}&$0.189$&$0.036$&$0.086$
    &$69.72_{\pm0.24}$ {\color{red}($-1.67$)}&$0.198$&$0.035$&$0.081$
    
    \\
    
    Greedy($\gamma_{0.6}$)
    &$93.54_{\pm0.24}${\color{green}($+0.17$)}&$0.236$&$0.039$&$0.105$
    &$71.73_{\pm0.48}$ {\color{green}($+0.06$)}&$0.266$&$0.036$&$0.109$
    
    \\
    
    Greedy($\gamma_{0.7}$)
    &$93.82_{\pm0.20}${\color{green}($+0.45$)}&$0.320$&$0.038$&$0.135$
    &$72.85_{\pm0.15}$ {\color{green}($+1.7$)}&$0.392$&$0.038$&$0.160$
    
    \\
    
    Greedy($\gamma_{0.8}$)
    &$94.54_{\pm0.18}${\color{green}($+1.17$)}&$0.725$&$0.045$&$0.285$
    &$75.61_{\pm0.29}$ {\color{green}($+4.57$)}&$0.925$&$0.045$&$0.360$
    
    \\
    
    Greedy($\gamma_{0.9}$)
    &$\mathbf{95.40_{\pm0.05}}$ {\color{green}($+2.03$)}&$4.224$&$0.097$&$1.544$
    &$\mathbf{79.18_{\pm0.20}}$ {\color{green}($+8.03$)}&$5.388$&$0.097$&$2.097$
    
    \\\hline

    SA($\gamma_{0}$)
    &$93.61_{\pm0.12}$ {\color{green}($+0.24$)}&$0.235$&$0.034$&$0.105$
    &$71.63_{\pm0.13}$ {\color{green}($+0.48$)}&$0.256$&$0.037$&$0.097$
    
    \\
    
    SA($\gamma_{0.5}$)
    &$93.69_{\pm0.13}$ {\color{green}($+0.32$)}&$0.247$&$0.036$&$0.115$
    &$71.70_{\pm0.01}$ {\color{green}($+0.55$)}&$0.297$&$0.037$&$0.111$
    
    \\
    
    SA($\gamma_{0.6}$)
    &$93.89_{\pm0.08}$ {\color{green}($+0.52$)}&$0.275$&$0.035$&$0.119$
    &$72.41_{\pm0.41}$ {\color{green}($+1.26$)}&$0.320$&$0.036$&$0.122$

    \\
    
    SA($\gamma_{0.7}$)
    &$94.22_{\pm0.19}$ {\color{green}($+0.85$)}&$0.437$&$0.038$&$0.189$
    &$74.02_{\pm0.64}$ {\color{green}($+2.87$)}&$0.526$&$0.037$&$0.181$
    
    \\
    
    SA($\gamma_{0.8}$)
    &$94.86_{\pm0.21}$ {\color{green}($+1.49$)}&$0.762$&$0.049$&$0.279$
    &$76.85_{\pm0.16}$ {\color{green}($+5.70$)}&$1.389$&$0.049$&$0.493$
    
    \\

    SA($\gamma_{0.9}$)
    &$\mathbf{95.76_{\pm0.06}}$ {\color{green}($+2.39$)}&$5.135$&$0.103$&$2.250$
    &$\mathbf{79.54_{\pm0.39}}${\color{green}($+8.39$)}&$5.597$&$0.103$&$2.241$
   \\\hlinewd{1pt}
		\end{tabular}
    }

\vspace{-1.2em}
\end{table*}

\subsection{Experimental Results}
We show a summary of the performance of our Greedy and Simulated Annealing channel search algorithms for ResNet34 \cite{he2015deep} and DARTS7 \cite{liu2019darts} in Table \ref{table_results}. Within the table, we show the results achieved with momentum scaling factor $\gamma \in \{0, 0.5, 0.6, 0.7, 0.8, 0.9\}$.
Both Greedy and SA algorithms were able to outperform the baselines in the selected models. SA yielded higher accuracy per parameter when compared to Greedy search as seen in Figure \ref{fig:ChannelEvolution}. For the same accuracy, SA was able to optimize to a much lower channel size in comparison with the greedy search. Specifically, SA achieved a parameter reduction for Greedy Search's best result on ResNet34 CIFAR100 by roughly 45.66\% in Table \ref{table_results}. For further validation, we introduce Figure \ref{fig:ResnetAccParam} showing the performance of our proposed algorithms compared to the baseline, compound scaling \cite{tan2020efficientnet}, and random scaling on Resnet34 CIFAR100. As seen, the fitted trend lines for Greedy and Simulated Annealing surpass that of compound scaling, random scaling, and outperform the baseline in terms of parameter count. \vspace{-1.2em}

\section{Conclusion}
In this work, we presented an efficient channel searching method for modulation of channel sizes within convolution neural networks. We introduced a novel metric, dubbed Quality Condition (QC), which we used to assess the performance of individual layers. We also introduced an automated dependency extraction algorithm, which represents a network as a Directed Acyclic Graph (DAG) and determines the dependent channel sizes that must be joint optimized. To reduce the computation load associated with searching iterations, we introduced Knowledge Distillation, a method that can transfer learnt weights of convolution layers between trials of changing channel sizes. The dependency extraction algorithm, metric, and knowledge distillation technique are fused into the channel size searching algorithm. The two variations of the searching algorithm, dubbed Greedy and Simulated Annealing, present direct and controlled stochasticity variations to optimization, respectively. Our approach has been shown to find optimal architectures which can out-perform baselines in accuracy by a large margin. Potential applications of this work include computationally constrained settings, such as in mobile systems, where efficient networks are favoured. We hope that this research can help accelerate future work in the field of deep learning by making networks more robust and efficient.

{
\bibliographystyle{IEEEtran}
\bibliography{bibliography}
}

\appendices

\begin{figure*}[t]
\label{Supplemental:visualizer-diagram}
\centering
 \includegraphics[width=\textwidth]{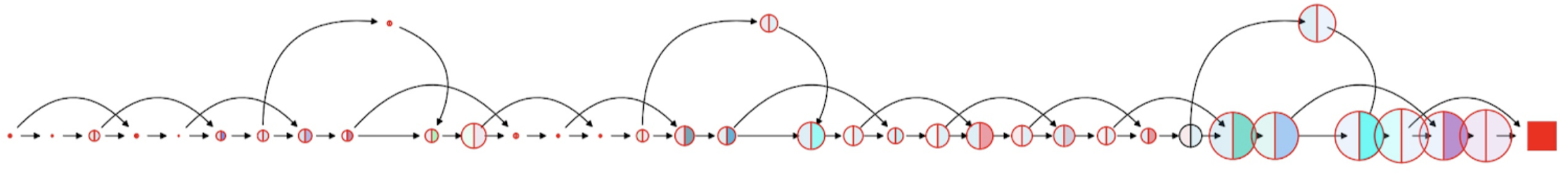}
  \begin{minipage}[t][0.6em][t]{1.0\linewidth}
  \caption{
  \textbf{Channel Size Visualizer.} Shows the Channel Size Visualizer of the Resnet34 Architecture generated automatically by our integrated pipeline - described in the main paper. The visualizer describes topology, channel size and inter-\textit{layer} dependencies of the individual Neural Architecture}
  \vspace{-1em}
\end{minipage}
 
\end{figure*}

\section{Channel Search Visualizer}
\label{supp:visualizer}
We introduce the Channel Size Visualizer as a tool used to probe intermediate layers of any CNN and develop a visual abstraction of it's structure including connections, channel size, and most notably, Channel Dependencies. Central to our paper has been the idea of Channel Size Optimization through the use of Algorithm 4 in the main paper. Figure \ref{Supplemental:visualizer-diagram} illustrates the Channel Size Visualizer applied to a standard ResNet34 Architecture. Each circle represents an individual \textit{Layer} - as defined in the main paper - with it's size corresponding to the channel size of that \textit{Layer}. The final square is in place to demonstrate the final class-determined layer in the Resnet34 Architecture - the Fully Connected Layer. Each circle is split into two equal sections representing the input and output channels of the respective \textit{Layer}. The color scheme of each \textit{Layer} is as follows: each unique dependency receives a color with each independent channel obtaining the defining color and each subsequent dependency obtaining a 40\% more transparent version of the said color. As a whole, this visualizer provides the user insight into the topology and intricacies of the neural network as to help understand the importance of channel dependencies towards the issue of algorithm design for channel size optimization.

\section{Ablative Studies}
\subsection{Effect of Alpha on Simulated Annealing.}
\begin{figure}[H]
\begin{minipage}{\columnwidth}
    
    \centerline{\includegraphics[width=0.95\columnwidth]{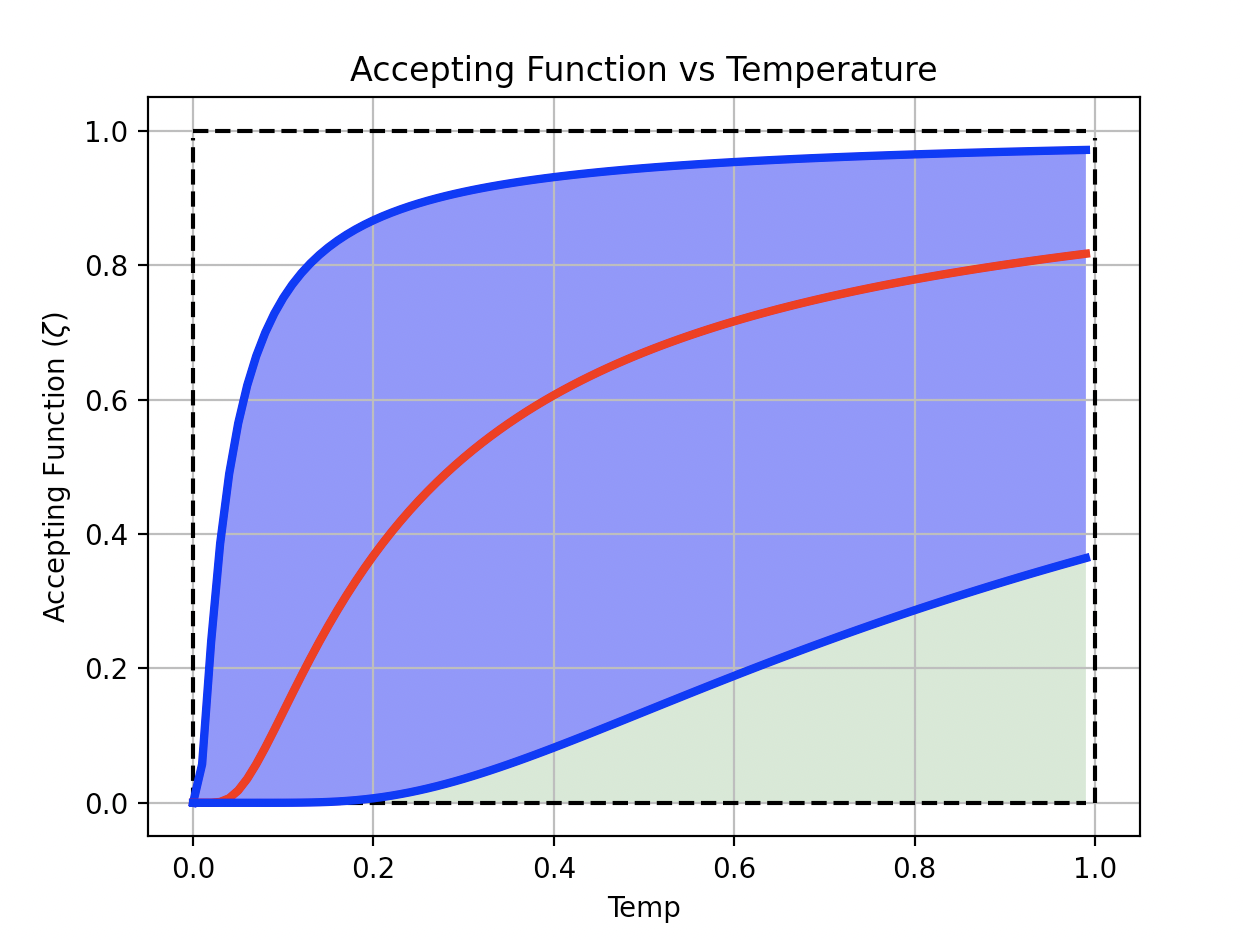}}
     \begin{minipage}[t][4.2em][t]{\linewidth}
    \vspace{-1.2em}
    \caption{\textbf{Accepting Function Alpha.} The dark blue lines represent $\alpha$ values of 35 and 1 from upper to lower respectively. The red curve represents the accepting function when $\alpha = 5$. A higher value of acceptance function means increased stochasticity while a lower value means decreased stochasticity. }
    \end{minipage}
    \label{fig:alpha_sweep}
    \end{minipage}
     \vspace{-1.2em}
\end{figure}
 In Figure \ref{fig:alpha_sweep}, we show the variations of different $\alpha$ values. The blue region represents a change in the value of the accepting function $\zeta$ from 35 to 1, from the upper boundary to lower boundary, respectively. A steeper curve means randomness is likely in a larger number of search trials, while a more gradual curve means that the likelihood for randomness is gradually decreased as trials progress. Because the accepting function directly affects the degree of stochasticity within Algorithm \ref{Algo:SA}, a small $\alpha$ value induces limited randomness, while a large value induces an excessive amount of randomness. Therefore, we settle with $\alpha=5$, as shown in red, which provides a good balance between random scaling, and metric based scaling of channel sizes.

 \subsection{Effect of Number of Search Trials.}
\begin{figure}[H]
\begin{minipage}{\columnwidth}
    \vspace{-1.4em}
    \centerline{\includegraphics[width=0.95\columnwidth]{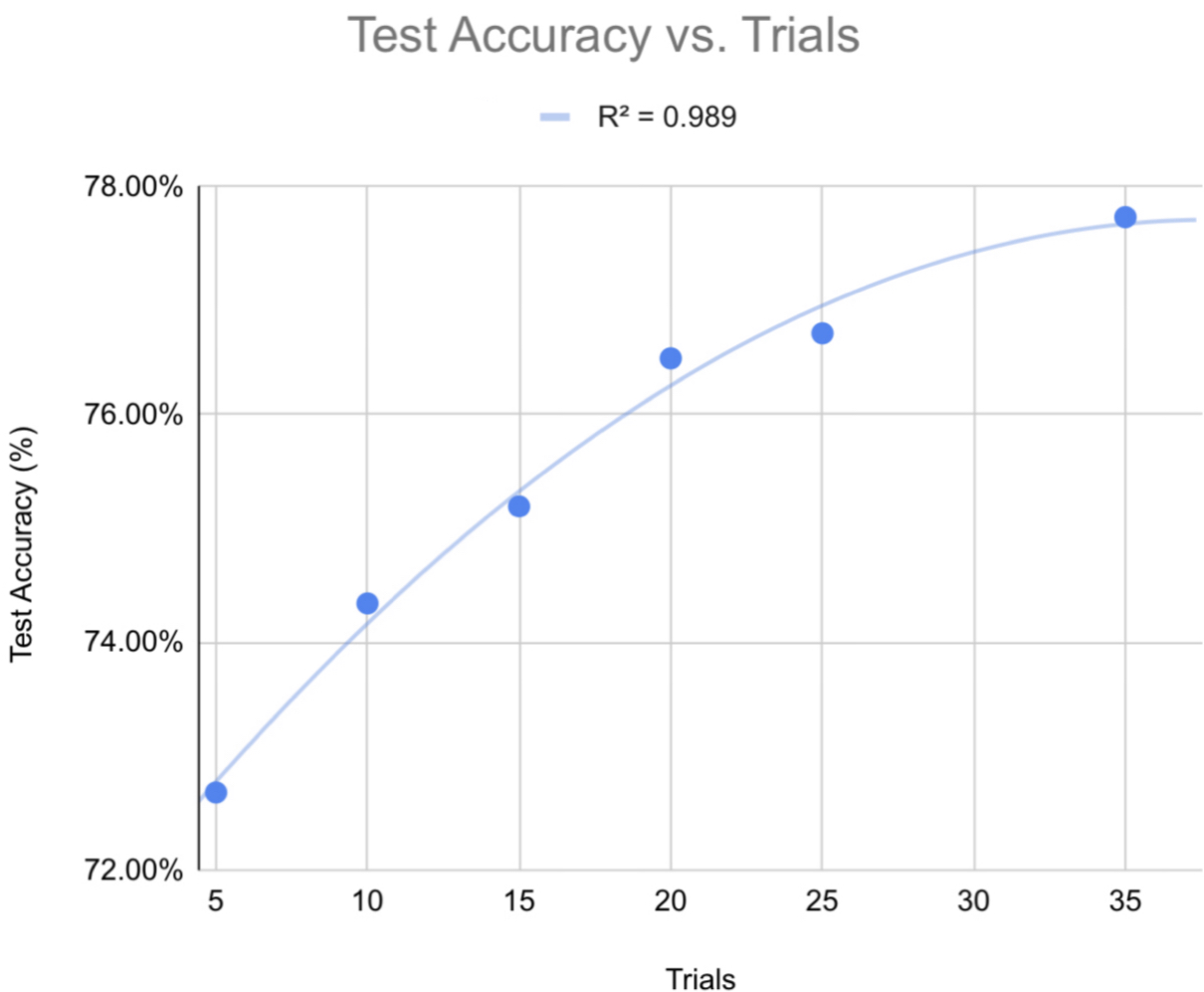}}
     \begin{minipage}[t][4.2em][t]{\linewidth}
    \vspace{-1.2em}
    \caption{\textbf{Effect of Trial Length.} We show the affect of Trial Length on final converged test accuracy (\%), and Parameter Count (M). We select the Trial Length of 35 for experimentation based on its balance between accuracy performance and computation cost.}
    \end{minipage}
    \label{fig:trial_sweep}
    \end{minipage}
     \vspace{-1.2em}
\end{figure}

 Figure \ref{fig:trial_sweep} illustrates the effect of different length search trials for channel size optimization using Greedy Search (Algorithm \ref{Algo:Greedy}). Since Greedy Search operates by choosing the best local optima, it makes for a good comparison in extrapolating how the length of the trials effects the accuracy of full train. This figure illustrates a sweep from 5 to 35 trials, and as seen by the correlation line, the trend plateaus at around 35 trials. This indicates that a further increase in trials beyond 35 would simply increase computational cost with a negligible accuracy gain. Since we focus on the micro-search space and finding a balance between accuracy and computational cost, we picked 35 trials as our selected hyper parameter for further experimentation.

\section{Evolution of Metrics Over Trials}
  \label{Metric-Evolution}
  
\begin{figure}[h]
\label{Supplemental:Greedy-metric-evol}
  \caption{\textbf{Metric Evolution of Greedy Algorithm.} We show the Metric Evolution of Greedy across 35 Trials. It can be seen that the metric values exhibit a period of randomness within the initial trials prior to stabilization. Upon stabilizing, the metrics for most layers exhibit an upwards trend until reaching a plateau in the later search trials.}
\minipage{0.5\columnwidth}
  \includegraphics[width=\linewidth]{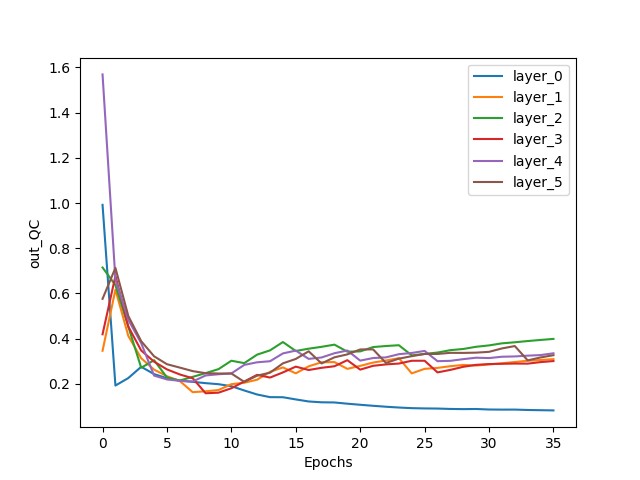}
\endminipage
\minipage{0.5\columnwidth}
  \includegraphics[width=\linewidth]{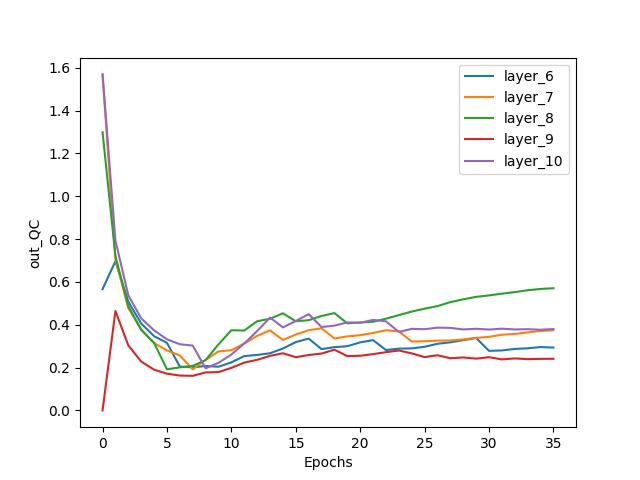}
\endminipage

\centering
\minipage{0.5\columnwidth}%
  \includegraphics[width=\linewidth]{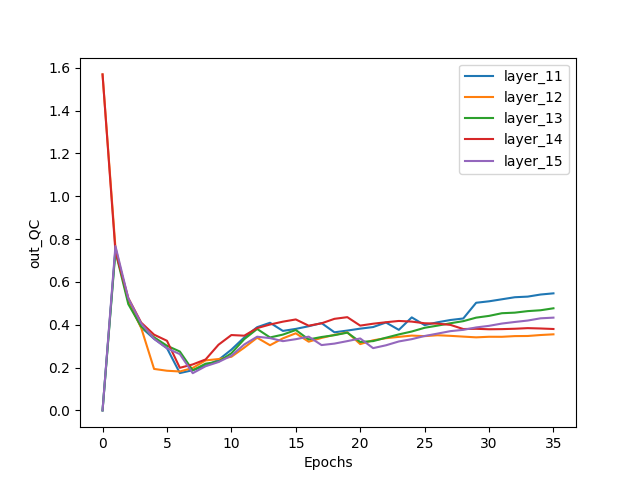}
\endminipage
\minipage{0.5\columnwidth}%
  \includegraphics[width=\linewidth]{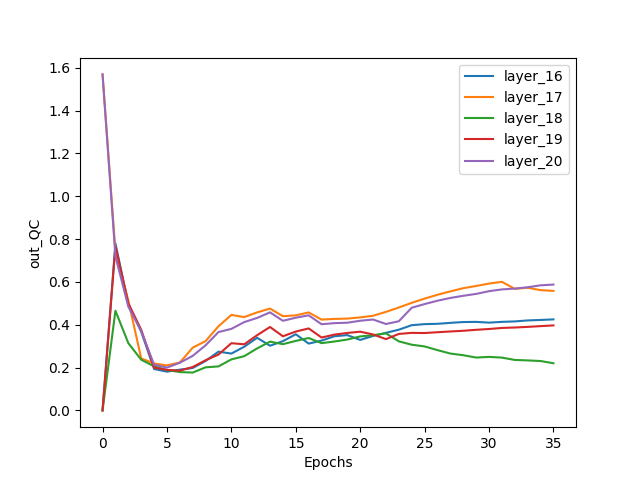}
\endminipage

\centering
\minipage{0.5\columnwidth}
  \includegraphics[width=\linewidth]{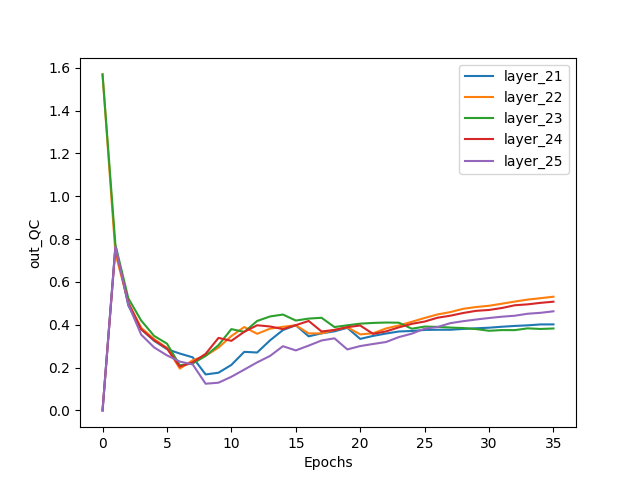}
\endminipage
\minipage{0.5\columnwidth}
  \includegraphics[width=\linewidth]{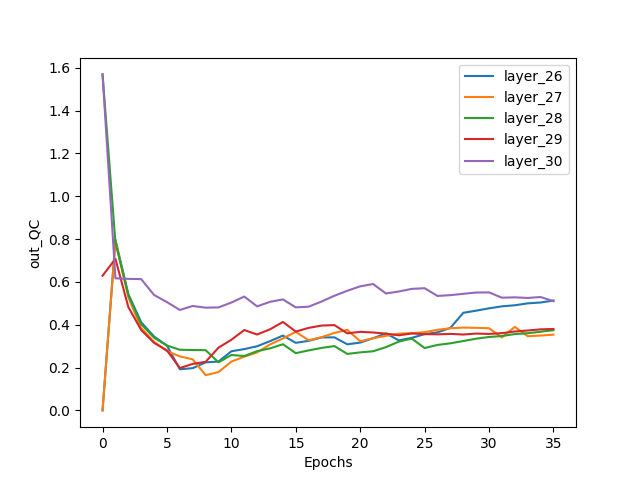}
\endminipage
\\
\centering
\minipage{0.5\columnwidth}%
  \includegraphics[width=\linewidth]{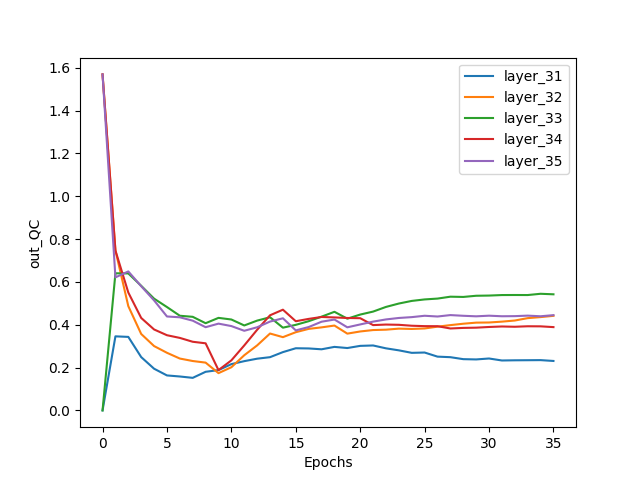}
\endminipage

\end{figure}

\begin{figure}[h]
\label{Supplemental:SA-metric-evol}
  \caption{\textbf{Metric Evolution of Simulated Annealing.} We show the Metric Evolution of Simulated Annealing across 35 Trials. It can be seen that the metric values exhibit a period of randomness within the initial trials prior to stabilization. Upon stabilizing, the metrics for most layers exhibit an upwards trend until reaching a plateau in the later search trials.}
\minipage{0.5\columnwidth}
  \includegraphics[width=\linewidth]{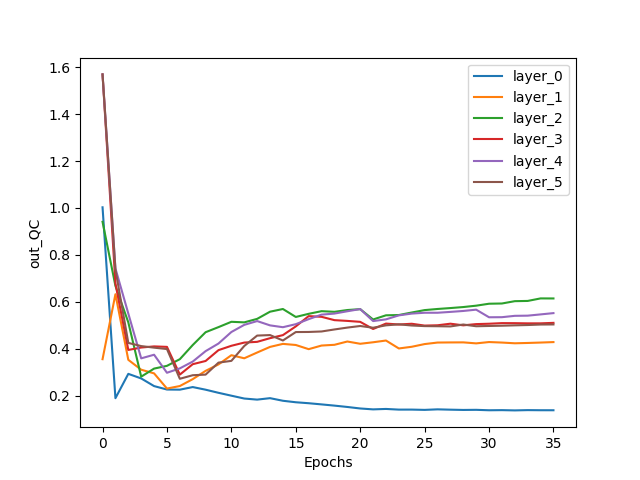}
\endminipage
\minipage{0.5\columnwidth}
  \includegraphics[width=\linewidth]{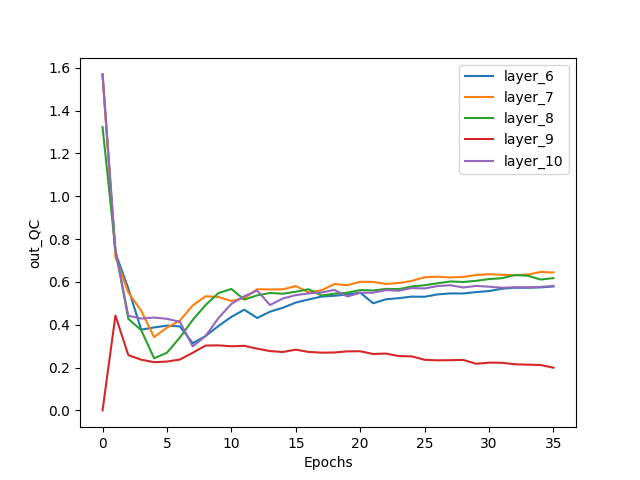}
\endminipage

\centering
\minipage{0.5\columnwidth}%
  \includegraphics[width=\linewidth]{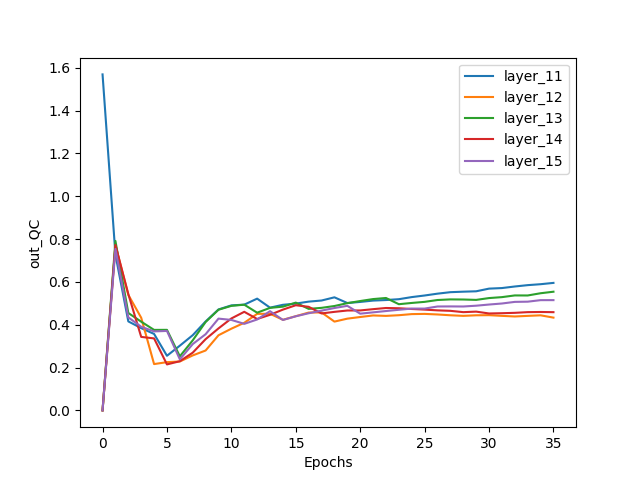}
\endminipage
\minipage{0.5\columnwidth}%
  \includegraphics[width=\linewidth]{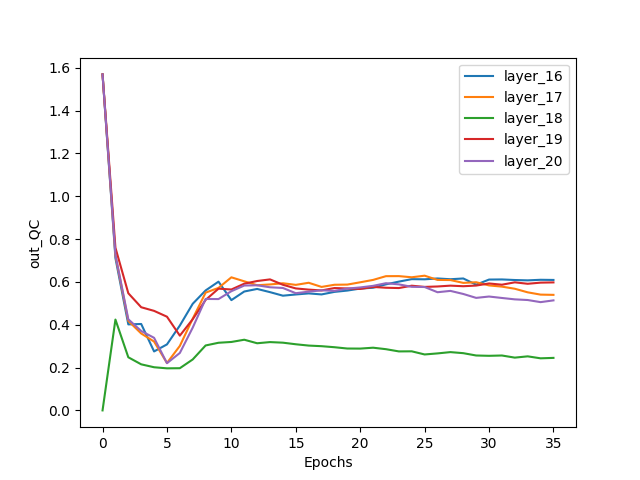}
\endminipage

\centering
\minipage{0.5\columnwidth}
  \includegraphics[width=\linewidth]{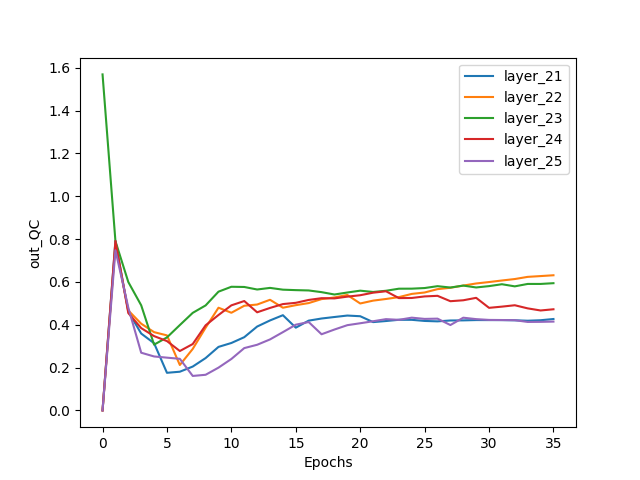}
\endminipage
\minipage{0.5\columnwidth}
  \includegraphics[width=\linewidth]{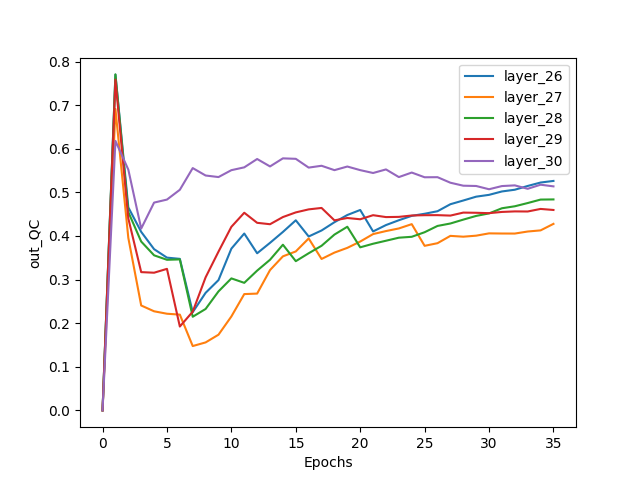}
\endminipage
\\
\centering
\minipage{0.5\columnwidth}%
  \includegraphics[width=\linewidth]{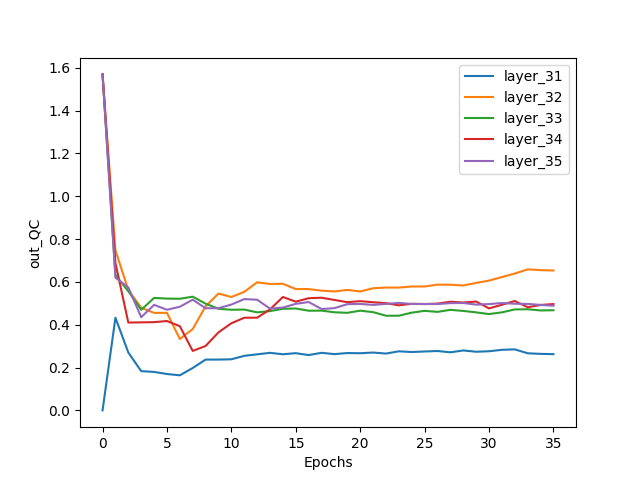}
\endminipage
\vspace{-1.5em}
\end{figure}

  In this section, we show the evolution of metrics of individual layers in ResNet34. Figure \ref{Supplemental:Greedy-metric-evol} presents the evolution of QC metric during searching using the Greedy Algorithm. Figure \ref{Supplemental:SA-metric-evol} presents the evolution of the QC metric for individual layers using the Simulated Annealing algorithm. it can be seen that for both algorithms, the layers experience instability in the initial trials, stablizes after the initial few trials and reaches a plateau in the latter half of the search trials. 
  
\clearpage
\vspace{5em}
\section{Cumulative Metric Evolution}
\label{supplemental:cumulative-metric}

\begin{figure}[h]
\label{Supplemental:cumulative-plot}
  \caption{\textbf{Cumulative Metric Evolution.} We show the evolution of cumulative QC metric for Greedy (left) and Simulated Annealing (right). We can see that the evolution of the cumulative metric for the greedy algorithm is more smooth than that of Simulated Annealing, due to the controlled stochasticity that is within the SA algorithm.}
\minipage{0.465\columnwidth}
  \includegraphics[width=\linewidth]{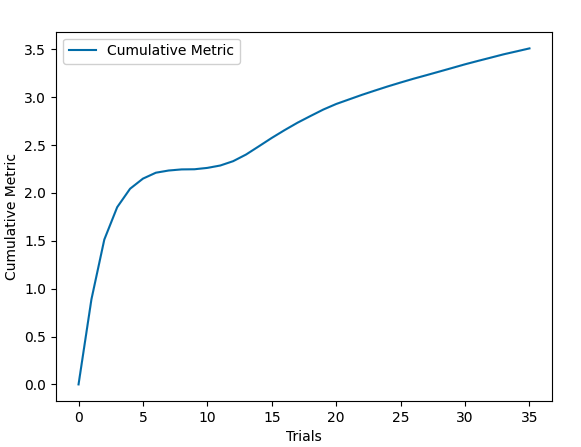}
\endminipage
\minipage{0.535\columnwidth}
  \includegraphics[width=\linewidth]{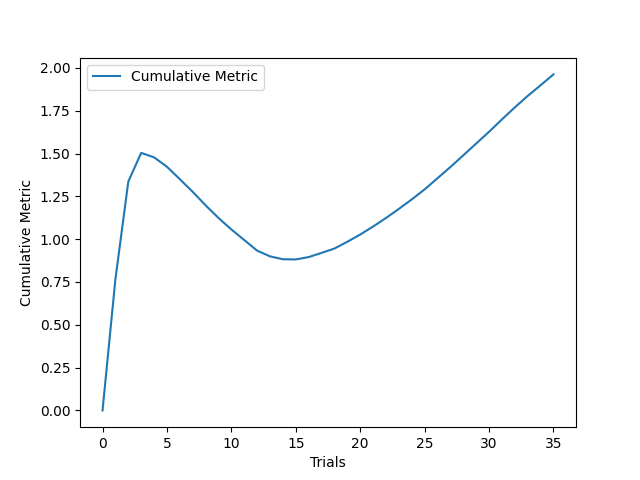}
\endminipage
\end{figure}

In this section, we show the evolution of the cumulative metric per trial for Greedy and Simulated Annealing in Figure \ref{Supplemental:cumulative-plot}. We compute the cumulative metric as follows:
\begin{equation}
    \mathcal{C} = \frac{1}{n_\mathcal{D}}\sum_{d \in \mathcal{D}} \sum_{l \in d} m_l
\end{equation}
Where $\mathcal{C}$ represents the value of the cumulative metric, $n_{\mathcal{D}}$ represent the total number of elements within the Dependency list $\mathcal{D}$ and dependency sublists $d$, $l$ represents the layers within the dependency sublist $d$, and $m_l$ represents the layer's metric.

It can be seen that the Greedy cumulative metric evolution is more smooth than that of Simulated Annealing. This is because the greedy algorithm takes a direct approach to always optimize for the best local heuristic. On the other hand, due to the stochasticity of Simulated Annealing, the graph takes detours to explore the seemingly "less optimal solutions" as deemed by Greedy.

\end{document}